\documentclass[twoside]{article}

\makeatletter
\usepackage[
    a4paper,
    twoside,
    bindingoffset=1cm,
    inner=1.5cm,
    outer=1.5cm,
    lines=60]{geometry}
\usepackage{fancyhdr}
\usepackage{ifthen}

\RequirePackage[T1]{fontenc}
\RequirePackage{amsmath,amsthm,amsfonts,amssymb}
\RequirePackage{txfonts}

\RequirePackage{graphicx}
\RequirePackage[font=small,labelfont=bf,
  labelsep=colon,singlelinecheck=false,aboveskip=3pt,justification=raggedright]{caption}

\AtBeginDocument{\twocolumn}
\renewcommand\maketitle{\par
  \begingroup
    \renewcommand\thefootnote{\@fnsymbol\c@footnote}\def\@makefnmark{\rlap{\@textsuperscript{\normalfont\@thefnmark}}}\long\def\@makefntext##1{\parindent 1em\noindent
            \hb@xt@1.8em{\hss\@textsuperscript{\normalfont\@thefnmark}}##1}\if@twocolumn
      \ifnum \col@number=\@ne
        \@maketitle
      \else
        \twocolumn[\@maketitle]\fi
    \else
      \newpage
      \global\@topnum\z@
      \@maketitle
    \fi
    \thispagestyle{fancy}\@thanks
  \endgroup
  \setcounter{footnote}{0}\global\let\thanks\relax
  \global\let\maketitle\relax
  \global\let\@maketitle\relax
  \global\let\@thanks\@empty
  \global\let\@author\@empty
  \global\let\@date\@empty
  \global\let\@title\@empty
  \global\let\title\relax
  \global\let\author\relax
  \global\let\date\relax
  \global\let\and\relax
}
\def\@maketitle{\newpage
    {\thispagestyle{fancy}\large\bfseries \noindent\@title \par}\vskip 1.5em{\lineskip .5em\noindent
        \@author
\par}\par
  \vskip 2em}

\renewcommand\section{\@startsection {section}{1}{\z@}{-3.5ex \@plus -1ex \@minus -.2ex}{2.3ex \@plus.2ex}{\normalfont\normalsize\bfseries}}
\renewcommand\subsection{\@startsection{subsection}{2}{\z@}{-3ex\@plus -1ex \@minus -.2ex}{1.25ex \@plus .2ex}{\normalfont\normalsize\bfseries}}
\renewcommand\subsubsection{\@startsection{subsubsection}{3}{\z@}{-3ex\@plus -1ex \@minus -.2ex}{1.25ex \@plus .2ex}{\normalfont\normalsize\bfseries}}

\def\@listi{\leftmargin\leftmargini
            \parsep 0\p@ \@plus2\p@
            \topsep 0\p@ \@plus2\p@
            \itemsep0\p@ \@plus2\p@}
\let\@listI\@listi

\renewcommand{\refname}{References}

\renewenvironment{abstract}
                {\setlength{\parindent}{0pt}\itshape}
                {\vspace{1\baselineskip}\par}

\RequirePackage{endnotes}
\let\footnote\endnote

\def\enoteformat{\rightskip\z@ \leftskip=1.5em \parindent=0em
   \leavevmode\llap{\makebox[1.5em]{\hfill\@theenmark.\hspace{.5em}}}}

\RequirePackage{hyperref}

\RequirePackage{breakurl}

\theoremstyle{plain}
\usepackage{aliascnt}

\newtheorem{theorem}{Theorem}

\newaliascnt{lemma}{theorem}
\newtheorem{lemma}[lemma]{Lemma}
\aliascntresetthe{lemma}

\newaliascnt{prop}{theorem}

\aliascntresetthe{prop}

\newaliascnt{property}{theorem}
\newtheorem{property}[theorem]{Property}
\aliascntresetthe{property}

\theoremstyle{definition}
\newaliascnt{definition}{theorem}
\newtheorem{definition}[definition]{Definition}
\aliascntresetthe{definition}

\theoremstyle{remark}
\newaliascnt{remark}{theorem}

\aliascntresetthe{remark}

\makeatother

\newcommand\myTitle{Optimization without Backpropagation}
\newcommand\myFullName{Gabriel Belouze}
\newcommand\myName{G. Belouze}
\newcommand\myUni{École Normale Supérieure de Paris}
\newcommand\myMail{gabriel.belouze@ens.psl.eu} \usepackage{maths}
\usepackage{acro}
\usepackage{xspace}
\usepackage{enumerate,enumitem}

\PassOptionsToPackage{backend=bibtex8,bibencoding=ascii,language=auto,style=numeric-comp,sorting=nyt, maxbibnames=10, natbib=true }{biblatex}
\usepackage{biblatex}

\usepackage{subfig}

\usepackage[cachedir=minted-cache,frozencache]{minted}
\newcommand{\wengert}{text}
\setminted[text]{escapeinside=~~,mathescape=true,breaklines=true}
\setminted[adsl]{escapeinside=~~,mathescape=true,breaklines=true}
\setminted[envision]{escapeinside=~~,mathescape=true,breaklines=true}
\setminted[python]{escapeinside=~~,mathescape=true,breaklines=true}
  \newsubfloat{listing}

\makeatletter
\renewcommand\p@sublisting{\thelisting}
\makeatother

\PassOptionsToPackage{fleqn}{amsmath} \usepackage{amsmath,empheq}
\usepackage{bm}

\usepackage{hyperref}
\newcommand\email[1]{\href{mailto:#1}{#1}}

\newcommand{\htarget}[1]{\hypertarget{#1}{#1}}
\newcommand{\hlink}[1]{{\hypersetup{hidelinks}\hyperlink{#1}{#1}}}

\makeatletter
\newcommand\Autoref[1]{\@first@ref#1,@}
\def\@throw@dot#1.#2@{#1}\def\@set@refname#1{\edef\@tmp{\getrefbykeydefault{#1}{anchor}{}}\xdef\@tmp{\expandafter\@throw@dot\@tmp.@}\ltx@IfUndefined{\@tmp autorefnameplural}{\def\@refname{\@nameuse{\@tmp autorefname}s}}{\def\@refname{\@nameuse{\@tmp autorefnameplural}}}}
\def\@first@ref#1,#2{\ifx#2@\autoref{#1}\let\@nextref\@gobble \else \@set@refname{#1}\@refname~\ref{#1}\let\@nextref\@next@ref \fi \@nextref#2}
\def\@next@ref#1,#2{\ifx#2@ and~\ref{#1}\let\@nextref\@gobble \else, \ref{#1}\fi \@nextref#2}
\makeatother

\makeatletter
\renewcommand\p@sublisting{\thelisting}
\makeatother

\usepackage[dvipsnames]{xcolor}
\definecolor{halfgray}{gray}{0.55} \definecolor{webgreen}{rgb}{0,.5,0}
\definecolor{webbrown}{rgb}{.6,0,0}
\hypersetup{
colorlinks=true, linktocpage=true, pdfstartpage=3, pdfstartview=FitV,
breaklinks=true, pdfpagemode=UseNone, pageanchor=true, pdfpagemode=UseOutlines,plainpages=false, bookmarksnumbered, bookmarksopen=true, bookmarksopenlevel=1,hypertexnames=true, pdfhighlight=/O,urlcolor=webbrown, linkcolor=RoyalBlue, citecolor=webgreen, pdftitle={\myTitle},
pdfauthor={\textcopyright\ \myName, \myUni},
pdfsubject={},
pdfkeywords={},
pdfcreator={pdfLaTeX},
pdfproducer={LaTeX with hyperref and classicthesis}
}

\title{\myTitle}
\author{\myFullName\thanks{\myUni\ (\email{\myMail})}}

\newcommand{\ie}{i.\,e.\xspace}

\newcommand{\eg}{e.\,g.\xspace}

\DeclareAcronym{ad}{
    short = AD,
    long = Automatic Differentiation
}
\DeclareAcronym{dp}{
    short = DP,
    long = Differentiable Programming
}
\DeclareAcronym{dsl}{
    short = DSL,
    long = Domain Specific Language
}

\DeclareAcronym{ir}{
    short = IR,
    long = Intermediate Representation
}
\DeclareAcronym{il}{
    short = IL,
    long = Intermediate Language
}
\DeclareAcronym{adsl}{
    short = ADSL,
    long = Automatically Differentiable Sub-Language
}
\DeclareAcronym{sgd}{
    short = SGD,
    long = Stochastic Gradient Descent
}
\DeclareAcronym{jvp}{
    short = jvp,
    long = Jacobian Vector Product
}
\DeclareAcronym{vjp}{
    short = vjp,
    long = Vector Jacobian Product
}

\bibliography{biblio}

\begin{document}

\maketitle

\begin{abstract}
\raggedright
Forward gradients have been recently introduced to bypass backpropagation in autodifferentiation, while retaining unbiased estimators of true gradients. We derive an optimality condition to obtain best approximating forward gradients, which leads us to mathematical insights that suggest optimization in high dimension is challenging with forward gradients. Our extensive experiments on test functions support this claim.
\end{abstract}

\section{Introduction}

\label{ch:introduction}

\paragraph*{Context}
\ac{ad} has become an ubiquitous tool for the machine learning practitioneer ; it enables one to use the Jacobian $J_f$ of a primal function $f : \R^n \rightarrow \R^m$, whilst only writing the program that computes the primal. This becomes crucial for any gradient based optimization schemes (or even higher order methods). It usually comes in two modes : Forward and Reverse. Forward mode is suited for problems where $n \ll m$ and only adds a constant factor of memory and time overhead. Reverse mode is suited for problems where $m \ll n$ and adds, in its vanilla form, an unbounded memory overhead factor. In many applications, $f$ is a loss, with $m=1, n \gg 1$, and reverse mode, despite its cost, is prefered.

However a recent paper \citetitle{baydin_gradients_2022} by \citeauthor{baydin_gradients_2022} \cite{baydin_gradients_2022} proposes a method to use forward mode AD even in cases where $m \ll n$, through the use of \emph{forward gradients}, which are unbiased estimates of the true gradients, and showed promising results in high dimensional contexts such as Deep Learning.

\paragraph*{Results} Among the family of acceptable forward gradients proposed by \citeauthor{baydin_gradients_2022}, we exhibit one that is optimal. We also find that the criterion for acceptability can be relaxed, and provide the associated relaxed family of optimal forward gradients. The form that takes optimal forward gradient is particularly simple and amenable to further analysis. This leads us to show that forward gradients have theoretical shortcomings in high dimensions.

We also further the experiments from \citep{baydin_gradients_2022} with both a much more comprehensive set of test functions, and set of optimizers, and find that in practice as well, we observe degrading performance of forward gradients in higher dimensions.

\paragraph{Outline} In \Autoref{ch:autodiff,ch:optimization}, we provide a primer on the theoretical background which our work relies upon.
\autoref{ch:autodiff} summarizes the state of the art of autodifferentiation. Specific attention is given to the difference between forward and reverse modes, and a detail presentation of \citetitle{baydin_gradients_2022} is given.
\autoref{ch:optimization} recalls classical gradient-based optimization algorithms in machine learning. Specifically, SGD and Adam (and its derivates) are reviewed, as they constitute the basis of our experiments.

In the second part, \Autoref{ch:optim-with-fwd-gradients,ch:experiments}, we detail our work and our results.
\autoref{ch:optim-with-fwd-gradients} is more theoretical by nature. There, we derive an optimal variation to \Citeauthor{baydin_gradients_2022}'s forward gradients, and provide analytical insights as to why forward gradient descent could be challenging.
Finally, \autoref{ch:experiments} presents our experimental results, and shows that forward gradients fail to match the results obtained from reverse autodifferentiation.

\autoref{ch:appendix} contains some complementary proofs.

\section{Automatic Differentiation}
\label{ch:autodiff}

This chapter acts as a primer on \acf{ad}. Automatic Differentiation is a family of algorithms that take a program that computes $f: \R^n \rightarrow \R^m$ and derive a program that computes the Jacobian $J_f: \R^n \rightarrow \R^m \times \R^n$. We review the two \emph{modes} of automatic differentiation, Forward and Reverse, and summarize specifically the article by \Citeauthor{baydin_gradients_2022} \citep{baydin_gradients_2022}, which offers a new perspective on the modes of differentiation and is seminal to this paper.

For a more comprehensive review of the theory on `autodiff', we recommend \Citeauthor{baydin_automatic_2018}\citep{baydin_automatic_2018}, and \Citeauthor{margossian_review_2019}\citep{margossian_review_2019} for details on implementation techniques, which we barely address.

\subsection{Automatic Differentiation in a nutshell} \label{sec:autodiff_forward_v_reverse}

\begin{figure}
\centering
\subfloat[Evaluation graph]{\label{fig:evaluation-graph}
    \includegraphics[width=0.45\linewidth]{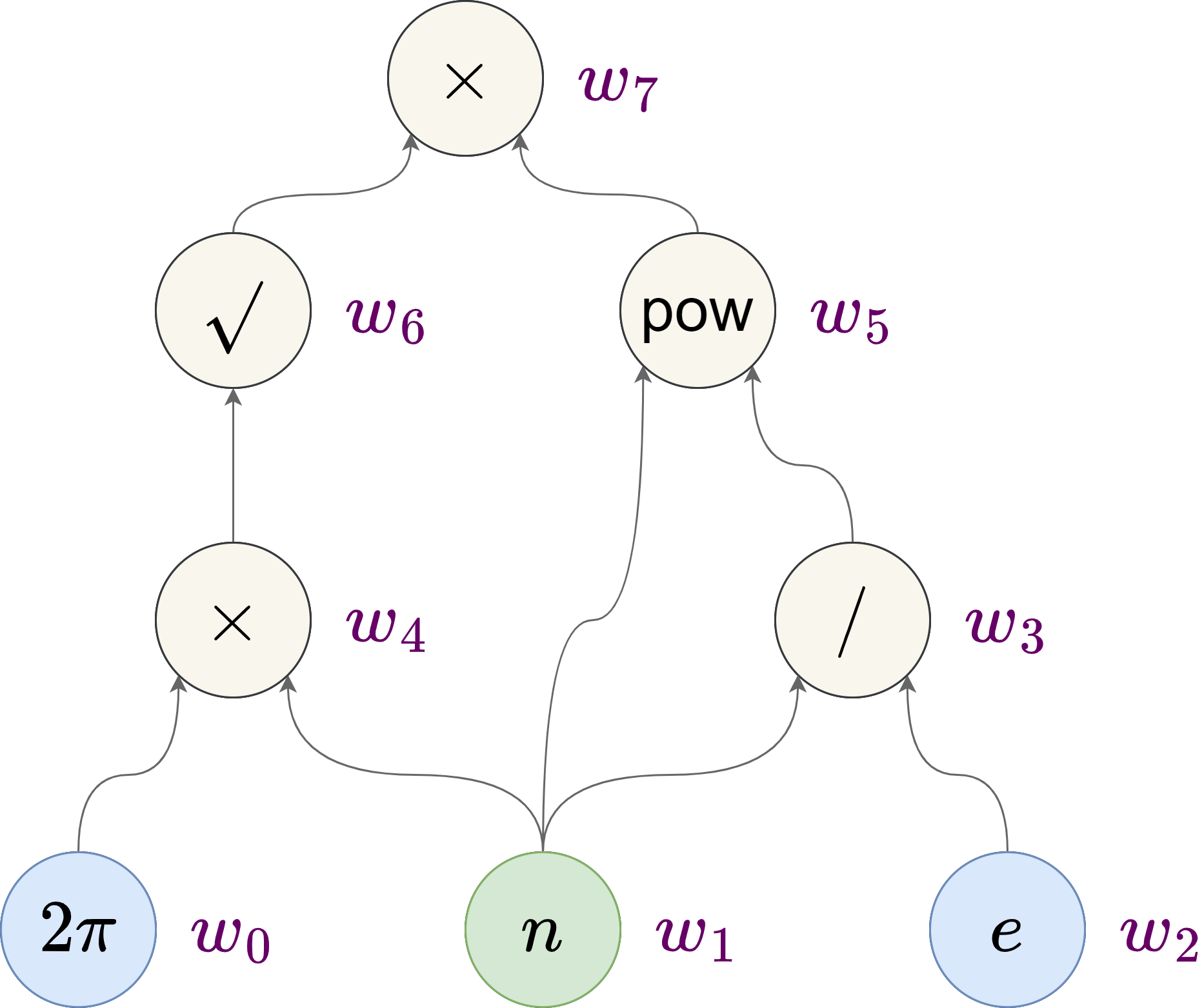}}\\
\subfloat[Wengert list]{
    \label{fig:wengert-list}
    \begin{minipage}[b]{0.35\linewidth}
\begin{minted}{\wengert}
w~$_0$~ <- 2~$\pi$~
w~$_1$~ <- ~$n$~
w~$_2$~ <- ~$e$~
w~$_3$~ <- w~$_1$~ / w~$_2$~
w~$_4$~ <- w~$_0$~ * w~$_1$~
w~$_5$~ <- w~$_3$~ ^ w~$_1$~
w~$_6$~ <- sqrt(w~$_4$~)
w~$_7$~ <- w~$_6$~ * w~$_5$~
\end{minted}
    \end{minipage}}
\caption{Representations of a program execution}
\label{fig:evaluation-graph-and-wengert}
\end{figure}

Take a program that computes some differentiable function $f: \R^n \rightarrow \R^m$, which may use complex control flow constructs, such as loops, conditionals or recursion. For a given evaluation point $\bm{x}$, we may record the computation flow to construct an \emph{evaluation graph}, which is a directed acyclic graph where nodes express an atomic computation (\autoref{fig:evaluation-graph}). This in turn may be flatten in topological sort order to ultimately obtain the \emph{evaluation trace} represented as a program in its simplest form, a Wengert list (\citeauthor{wengert_simple_1964} \citep{wengert_simple_1964}) (\autoref{fig:wengert-list}).

\paragraph*{}
At the heart of automatic differentiation is the chain rule. For each primal variable in the evaluation trace, we compute a differential variable which carries order 1 \emph{sensitivity} information. The specification for this information defines the \emph{mode} of the automatic differentiation. In both modes, the chain rules expresses relationships between those variables, which allows us to compute them.

\begin{description}[wide]
\item[Forward Mode] In the forward mode of automatic differentiation, we choose an \emph{initial tangent} $\bm{v} \in \R^n$. $v_i$ defines the sensitivity of the $i$-th parameter $x_i$. Then for each intermediary variable $w$ in the Wengert list, which mathematically correspond to some function $f_w$ of the input $\bm{x}$, we compute the \emph{tangent} of $w$, which is
$$\dot{w} \defeq J_{f_w} \cdot v$$
or less formally, $\dot{w} = \frac{\partial w}{\partial \x} \cdot \v$.
Consider now $w$ as an atomic function $\phi$ of $\bm{w}_{in} = (w_{i_1}, \ldots, w_{i_K})$, which is the vector of wengert variables that are the parent of $w$ in the computation graph. $\bm{w}_{in}$ is itself a function $f_{in}$ of the input, whence we get $f_w = \phi \circ f_{in}$. The chain rules then writes
$$J_{f_w}(\bm{x}) \cdot v = J_{\phi}(\bm{w}_{in}) \cdot J_{f_{in}}(\bm{x}) \cdot \bm{v} = J_{\phi}(\bm{w}_{in}) \cdot (\dot{w}_{i_1}, \ldots, \dot{w}_{i_k})$$
or again less formally, $\dot{w} = \sum_k \frac{\partial w}{\partial w_{i_k}} \cdot \dot{w}_{i_k}$. Thus, it is enough to know the jacobians of the atomic functions to inductively compute all tangents variable.\\
Ultimately, we compute simultaneously the primal $f(\x)$ and the \ac{jvp} $J_f(\x) \cdot \v$. This is a single pass of the forward mode -- if we wish to compute the full jacobian instead, then $n$ passes are necessary.
\item[Reverse Mode] In the reverse mode of automatic differentiation, we choose an \emph{initial cotangent} $\bm{u} \in \R^m$. For an intermediary Wengert variable $w$, we compute the \emph{adjoint} of $w$, which represents the sensitivity of the output $\bm{y} \in \R^m$ in the cotangent direction with respect to $w$. That is,
$$\bar{w} \defeq \frac{\partial \bm{y}}{\partial w} \cdot \bm{u}$$
Consider $\bm{w}_{out}=(w_{j_1}, \ldots, w_{j_L})$ the wengert variables that are children of $w$, with associated atomic functions \linebreak[4]{}$\phi_{j_1}, \ldots, \phi_{j_L}$. Now we can again leverage the chain rule to write
\begin{align*}
\bar{w} &= \frac{\partial \bm{y}}{\partial w} \cdot \bm{u} \\
&= \sum_l \frac{\partial w_{i_l}}{\partial w} \cdot \frac{\partial \bm{y}}{\partial w_{i_l}} \cdot \bm{u}\\
&= \sum_l \frac{\partial \phi_{i_l}}{\partial w} \cdot \bar{w}_{i_l}
\end{align*}
Again, it is enough to know the jacobians of the atomic functions to inductively compute all adjoints.\\
Ultimately, we compute the primal $f(\x)$ and the \ac{vjp} $J_f(\x)^T \cdot \u$. If we wish to compute the full jacobian instead, then $m$ passes are necessary.
\end{description}

There are two key differences between the two modes. First, if we wish to obtain the full jacobian, then the problem dimensions $n$ and $m$ will dictate which mode is better suited ; in particular in many machine learning contexts, $f$ is a loss and $m=1$, whence a single reverse mode pass is enough to obtain the full gradient. Second, the two modes compute sensitivity information in different orders : while forward mode compute the tangents in the same order as the primal, reverse mode starts from the end. This in fact is crucial for two reasons:
\begin{enumerate}[wide, labelwidth=!, labelindent=0pt]
\item Forward mode may be implemented by interweaving primal and tangent computations, and in particular does not require more than twice the primal memory usage. Reverse mode must be implemented in two passes, the forward -- or accumulation -- pass, and the infamous backpropagation pass. Before backpropagation, \emph{all} intermediary values must be stored. The memory overhead is proportionnal to the amount of computations, which in general is an unbounded factor of the primal memory bound (think for instance of \mintinline{text}{while} loops).
\item Forward mode does not require knowledge of the evaluation graph (the parents of a variable can be determined on the fly by looking at the atomic function call). Reverse mode must build the evaluation graph during the forward pass, which not only adds to the memory overhead, but also requires a more involved implementation.
\end{enumerate}

\subsection{Forward Gradient}

As stated earlier, forward mode autodifferentiation has better computational properties but is generally infeasible for many machine learning problems. We present here the idea by \citeauthor{baydin_gradients_2022}\citep{baydin_gradients_2022} which opens the door to forward mode for $f: \R^n \rightarrow \R$ even with $n \gg 1$.\\

Given an initial tangent, a single forward pass produces the \ac{jvp} $\langle \nabla f \cdot \v \rangle$, which tells us how much the gradient agrees with the tangent direction $\v$. Hence as a proxy for the true gradient, we can use the tangent scaled by this \ac{jvp}. That is, \Citeauthor{baydin_gradients_2022} define the \emph{forward gradient} to be $g(\theta) = \langle \nabla f \cdot \v \rangle \v$. The key idea is that this can be an unbiased estimate of the gradient granted $\v$ is sampled according to a carefully designed distribution. Let us first state the properties that this law must satisfy.

\begin{definition}[Tangent law properties]
\hypertarget{tangent law properties}{}We say that the probability law $\p$ on $\R^n$ satisfies the \emph{tangent law properties} when the marginals $(v_1, \ldots, v_n)$ of $\v \sim \p$ satisfy
\begin{subequations}\label{eq:tangent_law_conditions}
\begin{empheq}{align}
v_i \indep v_j \quad &\forall i \neq j\\
\esp(v_i) = 0 \quad &\forall i\\
\var(v_i) = 1 \quad &\forall i
\end{empheq}
\end{subequations}
\end{definition}

Now, the following theorem from \cite{baydin_gradients_2022} states that indeed the tangent law properties given above are enough to make the forward gradient a good estimator for the gradient.

\begin{theorem} \label{thm:forward_gradient}
Let $f: \R^n \rightarrow \R$, $\p$ that satisfies the \hlink{tangent law properties}, and $g$ the forward gradient associated to $f$ and $\p$. Then $g(\bm{\theta})$ is an unbiased estimator of $\nabla f(\bm{\theta})$.
\end{theorem}

Many convergence results, \eg for stochastic gradient descent, only assume given an unbiased estimate of the gradient, and not the true gradient. There are thus theoretical ground supporting the use of forward gradients. Furthermore, the authors conducted experiments with simple neural networks architectures, using $\mathcal{N}(\bm{0}_n, \bm{I_n})$ as tangent distribution, and report encouraging results : the network weights are optimized as fast as with true gradients in terms of epochs, and faster in terms of CPU time (this makes sense as forward mode is generally faster than reverse mode).

\section{Optimization in Machine Learning}

\label{ch:optimization}

Many machine learning problems ask to minimize an empirical risk objective, of the form
\[
    \mathcal{L}(\theta) = \frac{1}{M}\sum_{k=1}^M l(\theta; x_k)
\]
with respect to the vector parameter $\theta \in \R^n$. The dataset $\{x_1, \ldots, x_M\}$ being typically very large, gradient based optimization method use the batch gradient instead
\[
    g_B(\theta) = \frac{1}{|\mathcal{B}|} \sum_{k \in \mathcal{B}} \nabla l(\theta; x_k)
\]

where $\mathcal{B}$ is sampled uniformly in the subsets of $\{1, \ldots, M\}$ of size $B$. In the limit, $B$ is equal to $1$, and we obtain the simplest gradient scheme, \ac{sgd} \citep{robbins_stochastic_1951}, which performs the iterative updates

\[
    \theta_{t+1} = \theta_t - \alpha \cdot g_1(\theta_t)
\]

where the hyperparameter $\alpha$ is called the learning rate. When $\mathcal{L}$ is sufficiently regular, $g(\theta)$ being an unbiased estimator of the true gradient is enough to guarantee convergence, with rate $O(1/t)$ (see for instance \citeauthor{bach_lecture_2016}\citep{bach_lecture_2016}).

Other accelerated schemes are derived from vanilla SGD, notably its momentum variants (\citeauthor{nesterov_method_1983}\citep{nesterov_method_1983}), and may reach up to $O(1/t^2)$ convergence.

More recent methods, popular in the deep learning community, propose to update each coordinate of $\theta$ independently with still convergence guarantees (see \citeauthor{defossez_simple_2020}\citep{defossez_simple_2020}) ; among those, the ultra-widely-used Adam \citep{kingma_adam_2017}. Adam maintains an element-wise moving average of the gradients and of their square, called first and second moments.

\begin{align*}
\tilde{m}_{t+1,i} &= \beta_1 \tilde{m}_{t,i} + (1 - \beta_1) g(\theta_t)_i, \quad &m_{t,i} = \frac{\tilde{m}_{t,i}}{1 - \beta_1^{t+1}}\\
\tilde{v}_{t+1,i} &= \beta_2 \tilde{v}_{t,i} + (1 - \beta_2) g^2(\theta_{t})_i, \quad &v_{t,i} = \frac{\tilde{v}_{t,i}}{1 - \beta_2^{t+1}}\\
\end{align*}

The first moment acts as the usual gradient with heavy-ball momentum, and the second moment is used for element-wise scaling, yielding the following update

\[
    \theta_{t+1} = \theta_t - \alpha \cdot \frac{m_t}{\sqrt{v_t} + \varepsilon} \qquad \text{(operations are done element-wise)}
\]

As noticed by \citeauthor{balles_dissecting_2018}\citep{balles_dissecting_2018}, Adam can also be understood as a sign descent weighted inversely proportionnally to the relative variance of the gradient. That is

\begin{align*}
\frac{m_t}{\sqrt{v_t}} &= \frac{\text{sign}(m_t)}{\sqrt{v_t / m_t^2}} \\
    &= \text{sign}(m_t)\sqrt{\frac{1}{1 + \eta_t^2}}
\end{align*}

where $\eta_{t, i}^2 \defeq \frac{v_{t,i} - m_{t,i}^2}{m_{t,i}^2}$ is an approximation of the relative variance $\frac{\sigma_{t,i}^2}{\nabla \mathcal{L}_{t,i}^2}$, as long as $m_t$ and $v_t$ approximate well $\nabla \mathcal{L}$ and $\nabla \mathcal{L}^2$.

We will see later in \autoref{ch:optim-with-fwd-gradients} that this decoupling of Adam into those 2 aspects provides insights as to what using Adam with forward gradients amounts to. \\

Finally, we mention here Adabelief \citep{zhuang_adabelief_2020} as an alternative to Adam which is believed to be more stable to noisy gradients, and have better generalization properties than Adam. Adabelief is obtained by replacing the second moment of Adam with the moving average of empirical variance, \ie

\[
    \tilde{v}_{t+1} = \beta_2 \tilde{v}_t + (1 - \beta_2) (g(\theta_t) - m_t)^2
\]

The rational for this update is that $m_t$ can be interpreted as a prevision for the gradient, and $v_t$ as our confidence in the current gradient sample with respect to what the prevision was. As such, we take big steps when our confidence is high ($v_t$ is low), and conversely small steps when it is low.

\section{Optimization with Forward Gradients}
\label{ch:optim-with-fwd-gradients}

In this chapter, we study the mathematical implications underpinning the use of forward gradients, rather than true gradients, in standard optimization. \autoref{sec:choice-tangent-law} focuses on the choice  of distribution for the initial tangent ; we exhibit a \emph{best distribution} according to the constraints from \citeauthor{baydin_gradients_2022}, then show that those constraints can be relaxed, which notably yields a more general family of best distributions.  Finally, the important \autoref{sec:mistakes-forward-gradient} features evidence that optimization with forward gradient should get challenging with higher dimensional functions.
\subsection{Choice of Tangent Law}
\label{sec:choice-tangent-law}

\paragraph*{}Forward gradients are parametrized by a choice of direction of projection for the real gradient. To compute the forward gradient $g$ at $\bm{\theta}$, a random direction $\v$ is sampled and used as the \emph{tangent}. Forward mode AD then yields the jacobian vector product $\nabla f(\bm{\theta}) \cdot \v$, and finally the forward gradient is computed as $g(\bm{\theta}) \defeq (\nabla f(\bm{\theta}) \cdot \v) \v$. We naturally refer to the distribution for $\v$ as the \emph{tangent law}, and write $\p_{\v}$. From the \hlink{tangent law properties}, it is natural --though not necessary-- that the $v_i$ should be \iid, in which case we note the common distribution $\p_v$.

In the original forward gradient paper, \citeauthor{baydin_gradients_2022} use $\p_{\v} = \Normal(\mathbf{0}_n, \bm{I}_n)$, and note that any distribution that satisfies the \hlink{tangent law properties} yields valid forward gradients, in the sense that they are unbiased estimators of the gradient. This opens the door to other choices of tangent laws -- notably one that minimizes variance.

\begin{definition}[Minimal tangent law]
\hypertarget{minimal tangent law}{}We call \emph{minimal tangent law}, and we write $\p_v^{\min}$, the centered Rademacher law
\[ v \sim \Rad(0.5) \Leftrightarrow
    \begin{cases}
    v &= 1 \quad \textrm{\wp  $0.5$}\\
    v &= -1 \quad \textrm{\wp  $0.5$}
    \end{cases}
\]
\end{definition}

\begin{lemma}[Minimally deviating forward gradients] \leavevmode \\
\hypertarget{minimally deviating forward gradients}{}Let $f : \R^n \rightarrow \R$ be a function of $n$ real variables $(\theta_1, \ldots, \theta_n)$. Among the tangent laws that satisfy the \hlink{tangent law properties}, the choice $\p_v = \p_v^{\min}$ is \emph{the} minimizer for the forward gradient $g$ of the mean squared deviation $\esp\left [||g(\bm{\theta}) - \nabla f(\bm{\theta})||^2\right ]$.
\end{lemma}

\begin{proof}
After expansion of the mean squared deviation, this amounts to minimizing each $\esp\left[(g_i(\bm{\theta}) - \nabla f_i)^2\right]$. From the bias-variance property, and because the forward gradient is unbiased, this in turns asks to minimize $\var[g_i(\bm{\theta})]$.
The mean $\esp[g_i(\bm{\theta})]$ being fixed (equal to $\nabla f_i$), minimizing the variance amounts to minimizing each
\begin{align*}
\esp\left[g_i(\bm{\theta})^2\right] &= \esp\left[(\nabla f \cdot v)^2 v_i^2\right] \nonumber \\
&= \left(\frac{\partial f}{\partial \theta_{i}}\right)^2 \esp[v_i^4] + \sum_{j\neq i} \left(\frac{\partial f}{\partial \theta_j}\right)^2 \esp\left[v_i^2v_j^2\right] \nonumber \\
&\hphantom{=}+ 2\sum_{k < l} \left(\frac{\partial f}{\partial \theta_k}\right)\left(\frac{\partial f}{\partial \theta_l}\right) \esp\left[v_i^2v_kv_l\right] \\
&= \left(\frac{\partial f}{\partial \theta_{i}}\right)^2 (1 + \var[v_i^2]) + \sum_{j\neq i} \left(\frac{\partial f}{\partial \theta_j}\right)^2 \esp\left[v_i^2v_j^2\right] \nonumber \\
&= \left(\frac{\partial f}{\partial \theta_i}\right)^2 \var[v_i^2] + ||\nabla f||^2
\end{align*}

where we used the following properties
\begin{enumerate}
\item $\esp\left[v_i^2v_kv_l\right] = 0$ when $k \neq l$. Indeed the $v_j$ are independent and centered (from \autoref{eq:tangent_law_conditions}), and at least one of $v_k$, $v_l$ appears alone in $v_i^2v_kv_l$.
\item $\esp[v_i^2v_j^2]=1$ when $i \neq j$ (this follows again from \autoref{eq:tangent_law_conditions}).
\item $\esp[v_i^4] = 1 + \var[v_i^2]$.
\end{enumerate}

Finally, $\var[v_i^2]$ is minimum at $0$, which is feasible when $v_i$ takes value in $\{-a, a\}$ for some $a \in \R$. We obtain the \htarget{\emph{minimal deviation conditions}}

\begin{subequations}\label{eq:minimal_deviation_conditions}
\begin{empheq}[left=\empheqlbrace]{align}
v_i &\indep v_j &\forall i \neq j\\
\esp(v_i) &= 0 &\forall i\\
\var(v_i) &= 1 &\forall i\\
\var(v_i^2) &= 0 &\forall i
\end{empheq}
\end{subequations}

They are met only when the $v_i$ are independent Rademacher variables with parameter $0.5$, \ie  $\p_v = \p_v^{\min}$.
\end{proof}

Note that in the course of the proof we also found an explicit value for the mean squared deviation. The following property makes it explicit (see the \hyperlink{proof:minimal_deviation}{short proof} in the appendix).

\begin{property} \label{thm:minimal_deviation}
The mean squared deviation of the minimally deviating forward gradients is
\[\esp\left[\|\nabla f (\bm{\theta}) - g(\bm{\theta})\|^2\right] = (n-1)\left\|\nabla f(\bm{\theta})\right\|^2
\]
\end{property}

At this point, one may worry that the minimal tangent law that we found is anisotropic, while it is not obvious where the loss of isotropy happened. It is in fact the \hlink{tangent law properties} that implicitly assumes the anisotropic choice of canonical basis. Indeed, if the marginals of $\v$ according to some orthogonal basis $\b$ respect \autoref{eq:tangent_law_conditions}, the marginals according to some other orthogonal basis $\b'$ may not.

The \hlink{tangent law properties} can however be readily relaxed to an isotropic formulation. This the purpose of the following definition, and its associated theorem which extends \autoref{thm:forward_gradient}.

\begin{definition}[Extended tangent law properties]
\label{def:extented-tangent-law}
\hypertarget{extended tangent law properties}{}We say that the random vector $\w$ on $\R^n$ satisfies the \emph{extended tangent law properties} when there exists a random variable $\q \in \O_n$ over the orthogonal group of $\R^n$, and $\v$ that satisfies \autoref{eq:tangent_law_conditions}, such that $\w = \q \cdot \v$.
\end{definition}

Does the extended tangent law property add any useful distribution to the set of available tangent laws ? It arguably does. For instance, the uniform distibution over the $L_2$-sphere of radius $\sqrt{n}$, which is not admissible in the formulation of \citeauthor{baydin_gradients_2022}, satisfies the extended tangent law property. Indeed, it can be seen as the law of $\q \cdot \v$ where $\q$ is uniform over $\O_n$ (\ie following the translation invariant measure, or Haar measure, over the orthogonal group), and $\v$ follows the minimal tangent law.

\begin{theorem}[Extended forward gradient theorem]
\hypertarget{extended forward gradient theorem}{}Let $f: \R^n \rightarrow \R$, $\p$ that satisfies the \hlink{extended tangent law properties}, and $g$ the forward gradient associated to $f$ and $\p$. Then $g(\bm{\theta})$ is an unbiased estimator of $\nabla f(\bm{\theta})$.
\end{theorem}

\begin{proof}
Let $\w = \q \cdot \v$ be the random tangent associated to $\p$ (see \autoref{def:extented-tangent-law}). We rely on the decomposition $$\esp[g(\bm{\theta})] = \int_{q \in \O_n} \esp[g(\bm{\theta}) \; | \; \q=q] dq$$\\
Hence it suffices to show that $g(\bm{\theta}) $ is an unbiased estimator of the gradient given $\q$. We can simply recycle the proof of \autoref{thm:forward_gradient} from \citetitle{baydin_gradients_2022}, with a change of basis defined by $q$. We have
\begin{align*}
\langle \nabla f ; \w \rangle &= \langle q^{-1} \cdot \nabla f ; \v \rangle\\
\langle \nabla f ; \w \rangle \w &= q \cdot \langle q^{-1} \cdot \nabla f ; \v \rangle \v\\
\end{align*}
whence from \autoref{thm:forward_gradient} we get that $q^{-1} g(\bm{\theta}) $ is an unbiased estimator of $q^{-1} \nabla f(\bm{\theta})$, \ie that $g(\bm{\theta})$ is an unbiased estimator of $\nabla f(\bm{\theta})$. This concludes the proof that $\esp[g(\bm{\theta}) \; | \; \q=q] = \nabla f(\bm{\theta})$, and hence proves the theorem.
\end{proof}

Of course, the extended formulation yields an extended family of minimally deviating forward gradients.

\begin{definition}[Extended minimal tangent laws]
\hypertarget{extended minimal tangent laws}{}Let $\O_n$ be the set of orthogonal automorphism of $\R_n$. The family of \emph{extended minimal tangent laws} is described by the random variables $T = \q \cdot \v$ where $\v \sim \p_v^{\min}$ and $\q$ is any random variable over $\O_n$.
\end{definition}

\begin{figure}[h!]
\centering
\subfloat[Canonical minimal tangent law.]
{\includegraphics[width=.45\linewidth]{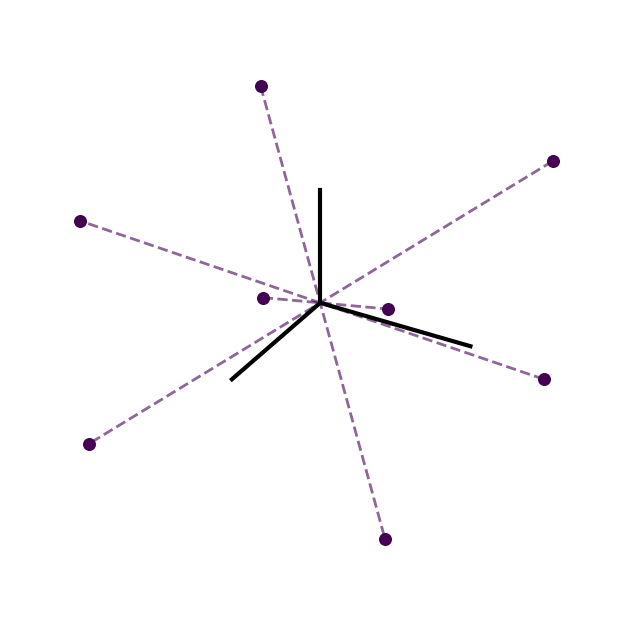}} \quad
\subfloat[Another minimal tangent law. Canonical basis is in black, rotated basis is in turquoise.]
{\includegraphics[width=.45\linewidth]{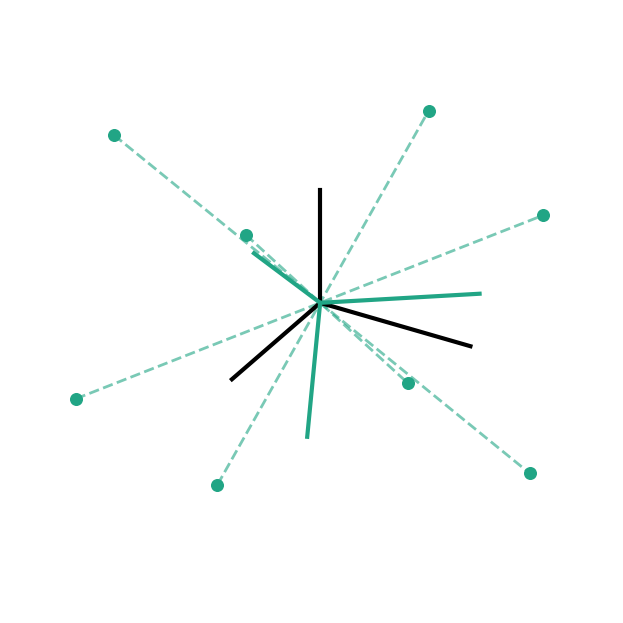}}
\caption{Minimally deviating tangent laws in dimension $n=3$. Each vertex has equal probability $1/8$.}\label{fig:minimal_tangent_laws}
\end{figure}

The simplest members of this family are those associated to almost surely constant $\q$, which are the independent centered Rademacher marginals associated to each orthogonal basis -- see \autoref{fig:minimal_tangent_laws}. The uniform distibution over the $L_2$-sphere of radius $\sqrt{n}$ also belongs to the extended minimal tangent laws, and is \emph{the} minimal tangent law that is isotropic.

We can furthermore check that all laws in this family are equivalently good (regardless of $\nabla f$), as the mean squared deviation given by \autoref{thm:minimal_deviation} is invariant to orthogonal transformations of $\v$.

\subsection{Mistakes in Forward Gradient Descent}
\label{sec:mistakes-forward-gradient}

\paragraph*{}Proofs of convergence for optimizers generally only assume an unbiased oracle of the gradient of the objective (see for instance \citeauthor{moulines_non-asymptotic_2011}\cite{moulines_non-asymptotic_2011} or \citeauthor{defossez_simple_2020}\cite{defossez_simple_2020}) -- usually to account for stochastic data samples. In particular, this means that we keep the same theoretical guarantees while using forward gradients. However, this does not account for the intuition that noisier oracles yield harder optimization. The goal of this section is to provide a \emph{quantitative description} of the added stochasticity that comes from using forward gradients rather than true gradients.

\paragraph*{}In the following, we only look at \hlink{minimally deviating forward gradients}. Not only is the \hlink{minimal tangent law} more convenient to analyze because of its simple form, but we expect them to yield better optimizations than any other tangent laws.

\subsubsection{The Curse of Dimensionality} \label{sec:curse_of_dimensions}

\paragraph{} The purpose of this section is to prove \autoref{thm:fwd_gradient_number_of_misdirections}, which shows that the sign of the forward gradient correlates with the sign of the true gradient only on $o(n)$ dimensions more than what pure chance accounts for.

\paragraph{} First, however, we provide two reasons we wish the sign of forward gradients to approximate well the sign of the true gradient. They both stem from the remark that forward gradients associated to a minimally deviating tangent law have a distinct structure, namely all their coefficients share the same magnitude. What consequence does this structure have with different optimizers ?

Consider first the case of Clipped SGD. Clipped SGD is a regularized version of SGD, which enforces an upperbound on the magnitude of the coordinates: in lieu of the gradient $\nabla f(\theta)$, clipped SGD uses its projection onto the ball $\{ \theta \; | \; \left\| \theta \right\|_{\infty} \leq 1\|\}$. That is, for coefficients that have a magnitude greater than 1, their sign is used instead. When using Clipped SGD with forward gradients, either all coefficients are transformed into their sign, or none are. Thus, when the gradient is large enough, all coefficients have too large a magnitude, Clipped SGD with forward gradients amouts to sign descent.

A similar mechanism can be seen with Adam. Recall from \autoref{ch:optimization} the decoupled interpretation of Adam, as coordinate-wise weighted sign descent. With forward gradients, the weights are in fact equal across coordinates, whence, up to this scaling factor, the optimization process amounts to sign descent.

\begin{theorem} \label{thm:fwd_gradient_number_of_misdirections}
The expected number of dimensions where the \hlink{minimally deviating forward gradients} has the same sign as the true gradient is upper bounded by
\[\frac{n}{2} + \sqrt{\frac{n}{2\pi}} + O(\frac{1}{\sqrt{n}})\]
\end{theorem}

\begin{proof}
Let us notice first that the forward gradient $g = (\nabla f \cdot v)v$ always has a positive correlation with the true gradient:
\[\nabla f \cdot g = (\nabla f \cdot v)^2 \geq 0\]
In other words, the forward gradient changes the sign of $\v$ if necessary, so that it correlates with $\nabla f$ (it also adds a scaling factor which is irrelevant for the theorem).

Now, first, let us assume that $\nabla f \in \{-1, 1\}^n$. Let us note $P$ the expected number of dimensions with the same sign as the true gradient, and $N$ that of dimensions with the opposite sign. We have
\begin{subequations}\label{eq:number_of_misdirections}
\begin{empheq}[left=\empheqlbrace]{align}
P + N &= n\\
\begin{split}
P - N &= \esp\left[ \sign(\sum_i \nabla f_i v_i) \sum_i \sign(\nabla f_i) v_i\right]\\
    &=\esp\left[ |\sum_i \nabla f_i v_i |\right]
\end{split}
\end{empheq}
\end{subequations}

By symmetry, $\esp\left[ |\sum_i \nabla f_i v_i|\right]$ is $\esp[|\sum_i v_i|]$. This is the well known problem of estimating how far away from $0$ does a $n$-step random walk goes. The derivation is detailed in \autoref{ch:appendix}, and gives $\esp[|\sum_i v_i|] = \sqrt{\frac{2n}{\pi}} + O(\frac{1}{\sqrt{n}})$. Eventually, we get
\[P = \frac{n}{2} + \frac{1}{2}\esp[|\sum_i v_i|] = \frac{n}{2} + \sqrt{\frac{n}{2\pi}} + O(\frac{1}{\sqrt{n}})\]

Remains to show that this holds as an upper bound for general $\nabla f$. This is readily obtained, as \autoref{eq:number_of_misdirections} now writes
\begin{empheq}[left=\empheqlbrace]{align*}
P + N &= n\\
P - N &= \esp\left[ \sign(\sum_i \nabla f_i v_i) \sum_i \sign(\nabla f_i) v_i\right]
\end{empheq}
where we have an upper bound
\[
\esp\left[ \sign(\sum_i \nabla f_i v_i) \sum_i \sign(\nabla f_i) v_i\right]\leq\esp\left[ |\sum_i \sign(\nabla f_i) v_i |\right]
\] where we previously had an equality.
\end{proof}

Note that a random walk in $\R^n$ would get right on average $\frac{n}{2}$ dimensions. Thus in high dimensions, \autoref{thm:fwd_gradient_number_of_misdirections} shows that forward gradients are only marginally better than pure randomness.

\subsubsection{Forward Gradient for Linear Objectives}

\paragraph*{}The essence of first-order gradient descent methonds is to use at $\bm{\theta}$ the best linear approximation of the objective and move accordingly. Here we provide analysis of the behaviour of forward gradients when the linear approximation is exact, \ie for a linear objective. Their ability -- or lack there of -- to degenerate the objective to $-\infty$ will give insights as to how it will fair against reverse mode gradients on more complex objectives.

In the following, we let $f(\bm{\theta}) \defeq \bm{\mu}^T \cdot \bm{\theta}$ for some $\bm{\mu} \in \R^n$.

\paragraph{SGD} \label{subsec:sgd-linear-objectives}

The update rule for vanilla $SGD$ writes here
\begin{empheq}[left=\empheqlbrace]{align*}
\d \bm{\theta}_t &= (\v_t^T \cdot \bm{\mu}) \v_t\\
\bm{\theta}_{t+1} &= \bm{\theta}_t - \d \bm{\theta}_t
\end{empheq}
where we assumed a learning rate of $1$ for simplicity, and where the $(\v_s)_{s=1,2,\ldots}$ are independent random variables with law $\p_{\v}^{\min}$.

We can compute the mean gain to the objective of a single SGD step
\begin{align*}
\esp\left[f(\bm{\theta}_{t+1}) - f(\bm{\theta}_t)\right] &= -\esp[(\v_t^T \cdot \bm{\mu})^2]\\
&= - \bm{\mu}^T \cdot \Sigma \cdot \bm{\mu}\\
&= - \|\bm{\mu}\|^2
\end{align*}
where the tangent covariance $\Sigma \defeq \esp[\v\cdot\v^T]$ is $\I_n$ (following the \hlink{tangent law properties}).

\begin{figure}[h!]
\centering
\subfloat[Evolution of the objective.]
{\label{fig:sgd_linear_objective__objective_evolution}\includegraphics[width=.45\linewidth]{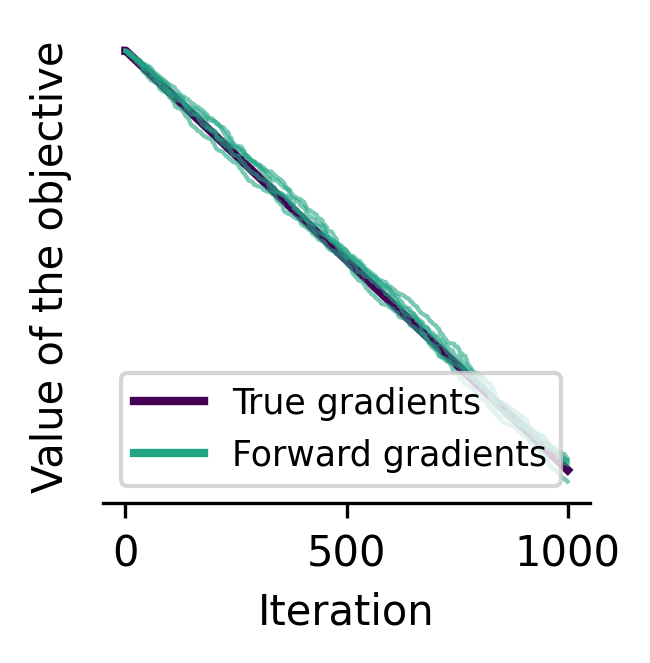}
} \quad
\subfloat[Evolution of the first two coordinates of $\bm{\theta}$]
{\label{fig:sgd_linear_objective__theta_evolution}\includegraphics[width=.45\linewidth]{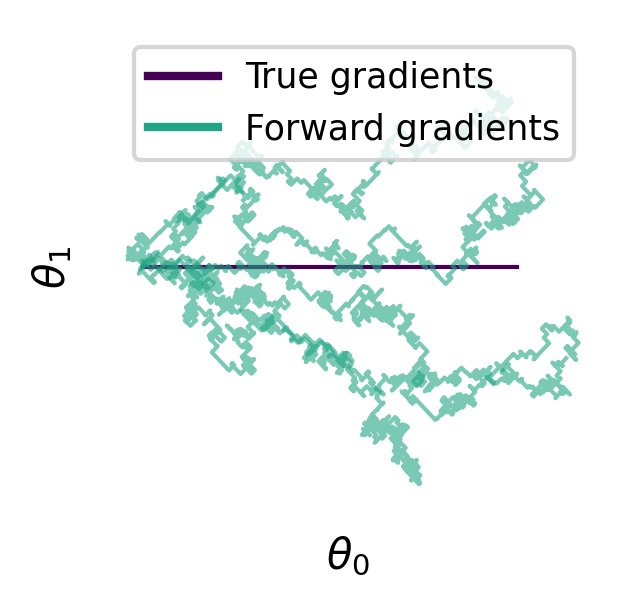}
}
\caption{SGD with linear objective, with forward gradients and true gradients. 5 iterations of forward gradient descent are represented.}\label{fig:sgd_linear_objective}
\end{figure}

This is the same expected gain as with regular reverse-mode SGD, and indeed \autoref{fig:sgd_linear_objective__objective_evolution} shows similar asymptotic evolution of the objective. Yet in high dimensions, we showed in \autoref{sec:curse_of_dimensions} that on average the forward gradient gets only $o(n)$ more than half directions correct. How does it manage to keep up with the always-correct true gradient ? Simply, forward SGD takes bigger steps.

Indeed consider the following simple derivation when $\v \sim \p_{\v}^{\min}$
\begin{align*}
\esp\left[\left\|(\nabla f^T \cdot \v)\v\right\|^2\right]&= n\esp\left[(\nabla f^T \cdot \v)^2\right] \\
&= n \|\nabla f\|^2\\
\end{align*}
Whereas SGD in reverse mode takes step of squared norm size $\|\nabla f\|^2$, forward adds on average a factor equal to the dimension. While this has no effect for the simple linear objective, this may slow down or prevent convergence on more chaotic problems. As seen in \autoref{fig:sgd_linear_objective__theta_evolution}, the parameter $\bm{\theta}$ may wander far from the gradient direction.

Note that in fact the derivation above remains valid if the gradient has only constant direction with possibly varying norm. That is, in a region where the isopleths are plane and parallel, we expect forward gradient descent and true gradient descent to perform equally well.

\paragraph{Adam}

Adam with constant gradient $\nabla f =\bm{\mu}$ amounts to sign descent (up to the Adam's $\varepsilon$), \ie $\delta \bm{\theta}_{i} = \sign(\bm{\mu}_i)$. The speed of divergence in this case is characterized by
\[\delta \bm{\theta}^T \cdot \bm{\mu} = \|\bm{\mu}\|_1\]

How does Adam with forward gradients compare ? The update rule for Adam writes as the following (we use $p_t$ for the second moment of Adam to not conflict with the notation for the initial forward tangent $\v$).

\begin{empheq}[left=\empheqlbrace]{align*}
m_t &= (1 - \beta_1)\sum_{s=1}^t \beta_1^{t-s} (\v_t^T \cdot \bm{\mu}) \v_t \\
p_t &= (1 - \beta_2)\sum_{s=1}^t \beta_2^{t-s} (\v_t^T \cdot \bm{\mu})^2 \bm{1} \\
\d \bm{\theta}_{t,i} &= \frac{\sqrt{1-\beta^t_2}}{1 - \beta^t_1} \frac{m_{t,i}}{\sqrt{p_{t,i}} + \varepsilon}\\
\bm{\theta}_{t+1} &= \bm{\theta}_t - \d \bm{\theta}_t
\end{empheq}
where we assumed a learning rate of $1$ for simplicity, and where the $\v_i$ are independent random variables with law $\p_{\v}^{\min}$. Typically, $\beta_1$ and $\beta_2$ are close to $1$. To derive analytical results, we will consider the degenerated case where they are both $1$, whence
\[
    \d \bm{\theta}_{t,i} = \frac{(\v_t^T \cdot \bm{\mu}) \v_{t,i}}{\sqrt{(\v_t^T \cdot \bm{\mu})^2}} = \sign(\v_t^T \cdot \bm{\mu}) \v_{t,i}
\]

(we also remove the $\varepsilon$ as we did in reverse mode).

Now we can compute the associated speed of divergence:
\begin{align*}
\esp[(\d\bm{\theta}_t \cdot \bm{\mu})] &= \esp[\sign(\v_t^T \cdot \bm{\mu}) (\v_t^T \cdot \bm{\mu})] \\
&= \esp[ \left | \v_t^T \cdot \bm{\mu} \right |]
\end{align*}

In all generality, we can only bound this coefficient by the one obtained with true gradients: $\esp[(\d\bm{\theta}_t \cdot \bm{\mu})] \leq \|\bm{\mu}\|_1$. However, for most $\bm{\mu}$, this bound is very coarse, and there is an additional factor of degradation proportional to the dimension.

In particular, when $\mu \in \{-1,1\}^n$, we get a $O(\sqrt{n})$ speed of divergence (see the derivation in \hyperlink{proof:minimal_deviation}{the appendix}), while reverse mode gives a $O(n)$ speed of divergence. We find here another `curse of dimension': as the dimension of the parameter grows, we expect forward gradients to perform increasingly worse than true gradients. \section{Experiments}
\label{ch:experiments}

The litterature provides an extensive body of \emph{test functions} meant to challenge and evaluate new optimization algorithms, see for instance \cite{kochenderfer_algorithms_2019} (Appendix B) and \cite{yang_test_2010} for a curated list of such function, or \cite{jamil_8_2013} for a comprehensive survey. \Citeauthor{baydin_gradients_2022} test in \cite{baydin_gradients_2022} two such functions -- Beale and Rosenbrock -- against forward gradients, with vanilla SGD, and show positive results. However both functions are 2-dimensional, which, according to our work in \autoref{sec:curse_of_dimensions}, should not make apparent the potential pitfalls of forward gradients.

Thus, this section is meant to provide additional `unit testing' experiments against forward gradients. It is not trivial to conduct a meaningful comparison experiment between optimization algorithms, that can draw generalizable conclusions. We followed the recommendations from, and refer the reader to, \citetitle{beiranvand_best_2017}\citep{beiranvand_best_2017}. In particular, we use diverse test functions and we test convergence from several starting points.

\subsection{Specification}

Here we detail our experimental protocol.

\paragraph{Functions used}

Out of reproducibility concerns, we brought Beale and Rosenbrock in our experiments. Note that Rosenbrock has a definition for arbitrary dimensions, although it was used only in dimension 2 in \cite{baydin_gradients_2022} (known then as the banana function).

For convergence plots, we use the simple sphere function, and the hyperellipsoid function, which are both separable. Separability ensures the problem should be trivial for simple gradient descent, while forward gradient descent may still struggle in high dimensions. The hyperellipsoid function is similar to the sphere function but is anisotropic. This may be relevant to test, considering the shape of the \hlink{minimal tangent law}.

\begin{align*}
f_{Sphere}(\x) &= \sum_{i=1}^D x_i^2\\
f_{Ellipsoid}(\x) &= \sum_{i=1}^D ix_i^2\\
\end{align*}

Our main concern remains robustness of results, which demands a large set of test functions. We used the collection implemented by \citeauthor{thevenot_optimization_2022}\citep{thevenot_optimization_2022}, which regroup 78 test functions well known in the litterature, and include convex, non-convex, separable, non-separable, multimodal and non-multimodal functions. \\

For each of those functions that are defined in arbitrary dimension, we test them in dimensions $2$, $10$ and $100$.

\paragraph{Optimizers used}

We chose to test SGD, Clipped SGD, Adam and Adabelief. SGD yields the most bare view of the behaviour of forward gradients. However SGD is sensitive to variations of the learning rate, and prone to diverge, and hence we also include Clipped SGD as a regularized version of SGD.
Adam is maybe the most popular optimizer in machine learning. Adabelief appears to exhibit better robustness to gradient noise than Adam, which is of course highly desirable with forward gradients.

For each such optimizer, we test a version that uses forward gradients, and one that uses true gradients.\\

We did not attempt any hyperparameter optimization scheme, and keep a fixed learning rate equal to $0.01$.

\paragraph{Reproducibility}

Our `unit testing' experiments were done in Python. Our implementation and experimentations are publicy available \href{https://github.com/gbelouze/forward-gradient}{on Github}, it relies on the \href{https://github.com/HIPS/autograd}{autograd} library \citep{maclaurin_autograd_2015} for performing autodifferentiation.

\subsection{Results}

Here, we are only interested in the performance of the optimizers with respect to the number of gradient evaluation. In particular, we do not compare the CPU time of execution, nor the memory usage as we make no attempt in our implementation to optimize one nor the other.\\

\paragraph*{Trajectory and convergence plots}
We first start by providing some trajectory and convergence plots. They are not the most practical, as they do not allow comparisons across test functions, but nevertheless can be useful to form an idea of how optimizers behave.

\begin{figure}[h!]
\centering
\subfloat[Beale trajectory plot.]
{\includegraphics[width=.45\linewidth]{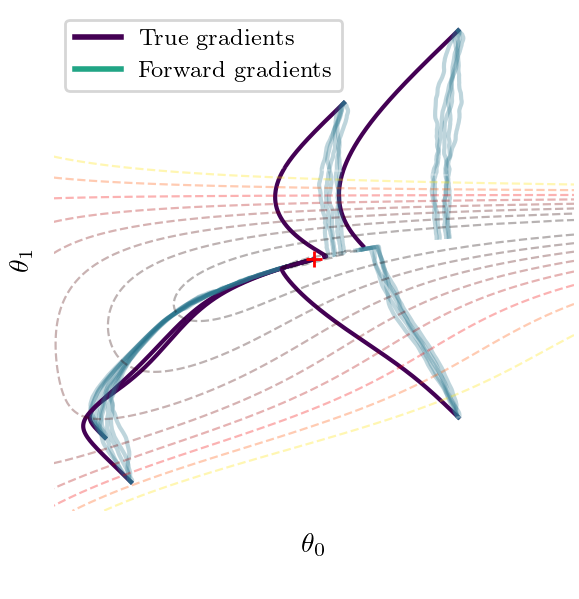}} \quad
\subfloat[Beale convergence plots for each initialization.]
{\includegraphics[width=.45\linewidth]{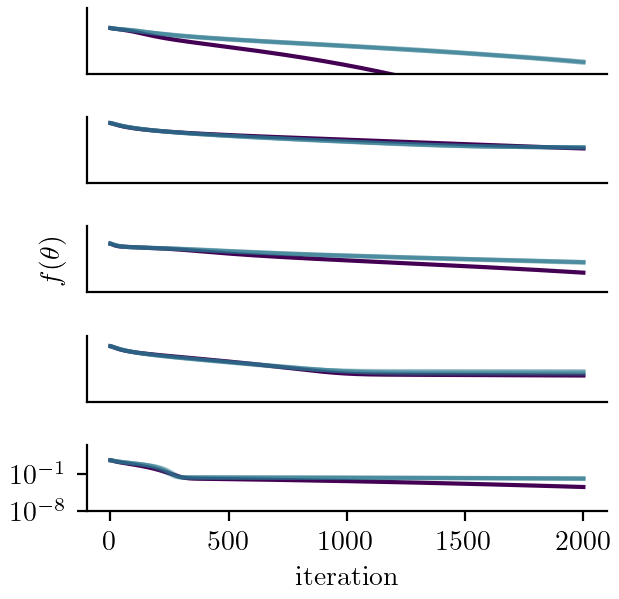}}
\\
\subfloat[Banana trajectory plot.]
{\includegraphics[width=.45\linewidth]{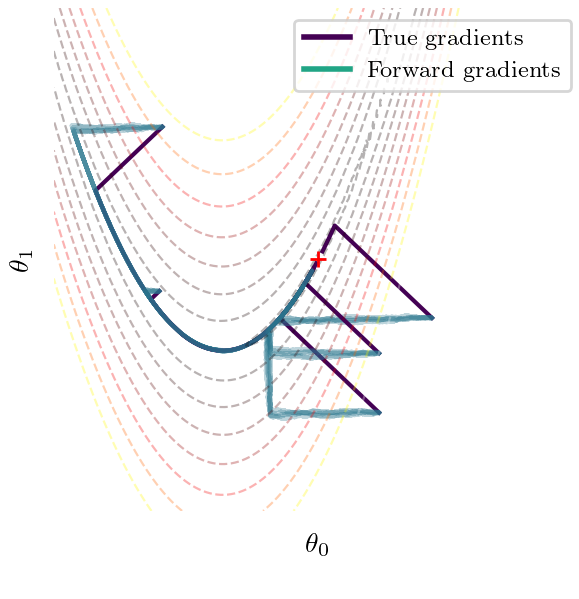}} \quad
\subfloat[Banana convergence plots for each initialization.]
{\includegraphics[width=.45\linewidth]{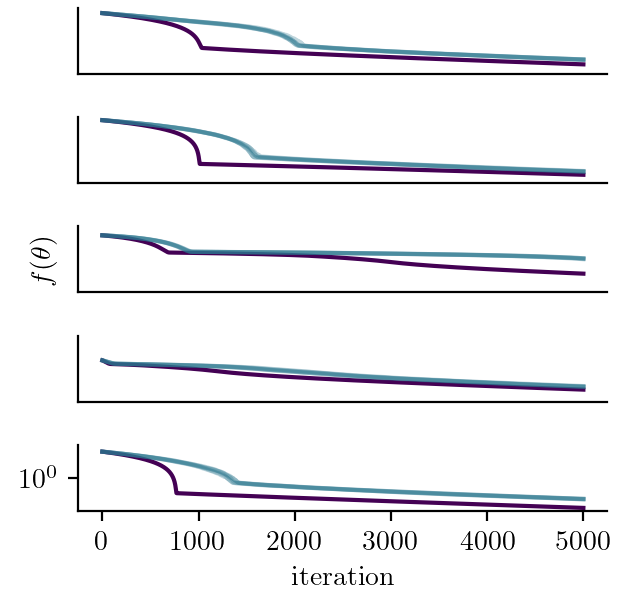}}
\caption{Vanilla gradient descent for the Beale and Banana test functions, with 5 random initializations.}
\label{fig:beale-and-rosenbrock}
\end{figure}

\autoref{fig:beale-and-rosenbrock} reproduces the results from \citeauthor{baydin_gradients_2022} with several initializations for $\theta$. It can be seen that it depends on the initialization wether true gradient descent or forward gradient descent performs better, although forward gradients are never but marginally better.

\begin{figure}[h!]
\centering
\subfloat[Dimension $2$]
{\includegraphics[width=.48\linewidth]{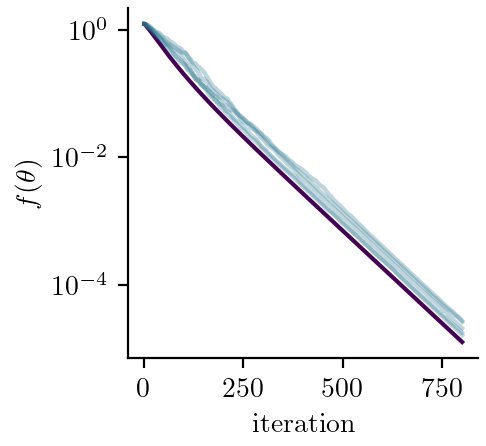}\hfill
\includegraphics[width=.48\linewidth]{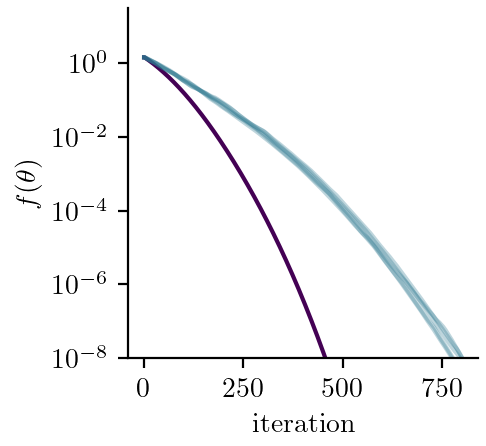}}\\
\subfloat[Dimension $10$]
{\includegraphics[width=.48\linewidth]{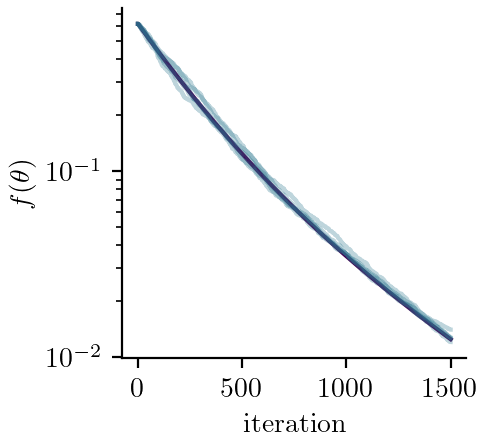}\hfill
\includegraphics[width=.48\linewidth]{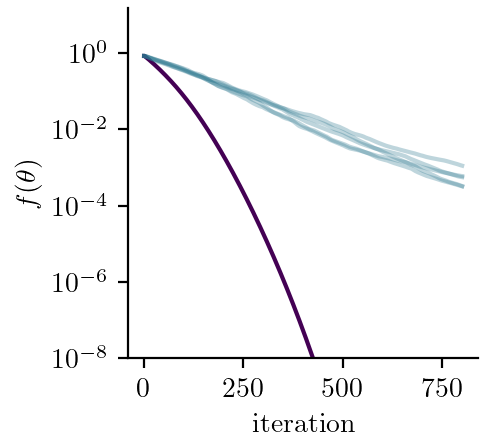}}\\
\subfloat[Dimension $100$]
{\includegraphics[width=.48\linewidth]{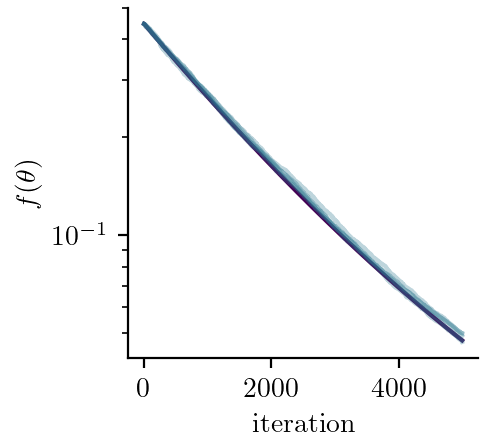}\hfill
\includegraphics[width=.48\linewidth]{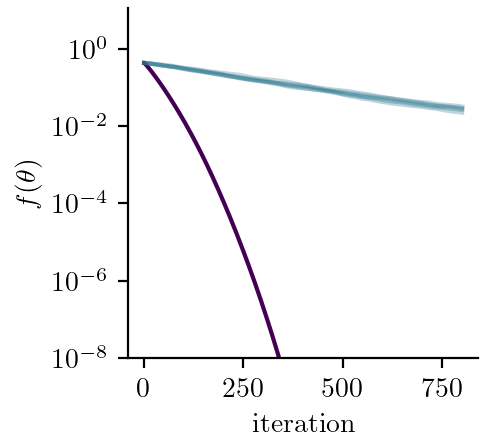}}
\caption{Evolution of SGD (left half) and Adam (right half) with the hyperellipsoid of dimension $2$, $10$ and $100$.}
\label{fig:hyperellipsoid-sgd-adam}
\end{figure}

We can furthermore see how the convergence plots evolve as the function dimension grows, which we illustrate with the ellipsoid function in \autoref{fig:hyperellipsoid-sgd-adam}. We can see that forward SGD does not seem to suffer from dimensionality. Indeed for low enough learning rates, the hyperellipsoid isopleths look straight and parallel on the path of gradient descent, which we showed in \autoref{subsec:sgd-linear-objectives} to be a case where forward gradients can readily replace true gradients. However, the same does not hold for Adam which displays convincingly how forward gradients struggle with high dimensionality.

\paragraph*{Performance profile} \citeauthor{beiranvand_best_2017}\citep{beiranvand_best_2017} recommend using performance profiles to report optimization experiments in a single graphic.

Let $\mathcal{P}$ be a set of problems (here test functions), and $\mathcal{S}$ a set of optimizers. Suppose our experiments produce a fixed-target metric $t_{p,s}$ for each problem $p$ and solver $s$, here we use the number of function evaluation to reach the minimum up to $\epsilon = 0.1$. Then we define the performance ratio
\[
    r_{p,s} = \begin{cases}
    \frac{t_{p,s}}{\min\{t_{p,s} \; | \; s \in \mathcal{S}\}} \quad&\text{if convergence test passed,}\\
    \infty &\text{otherwise}
    \end{cases}
\]
From this we derive the \emph{performance profile}, which is a function of $\tau \geq 1$ for each solver :
\[
    \rho_s(\tau) = \frac{1}{|\mathcal{P}|}\text{size}\{p \in \mathcal{P} \; | \; r_{p,s} \leq \tau \}
\]
That is, $\rho_s(\tau)$ is the proportion of problems where solver $s$ is less than a factor $\tau$ away from the best solver.

In particular, $\rho_s(1)$ is the portion of time that $s$ was the best solver, and $\rho(\infty)$ is the proportion that the solver managed to solve. In general, we are looking for solvers with consistently high $\rho$.\\

\begin{figure*}[h!]
\centering
    \includegraphics[width=.98\linewidth]{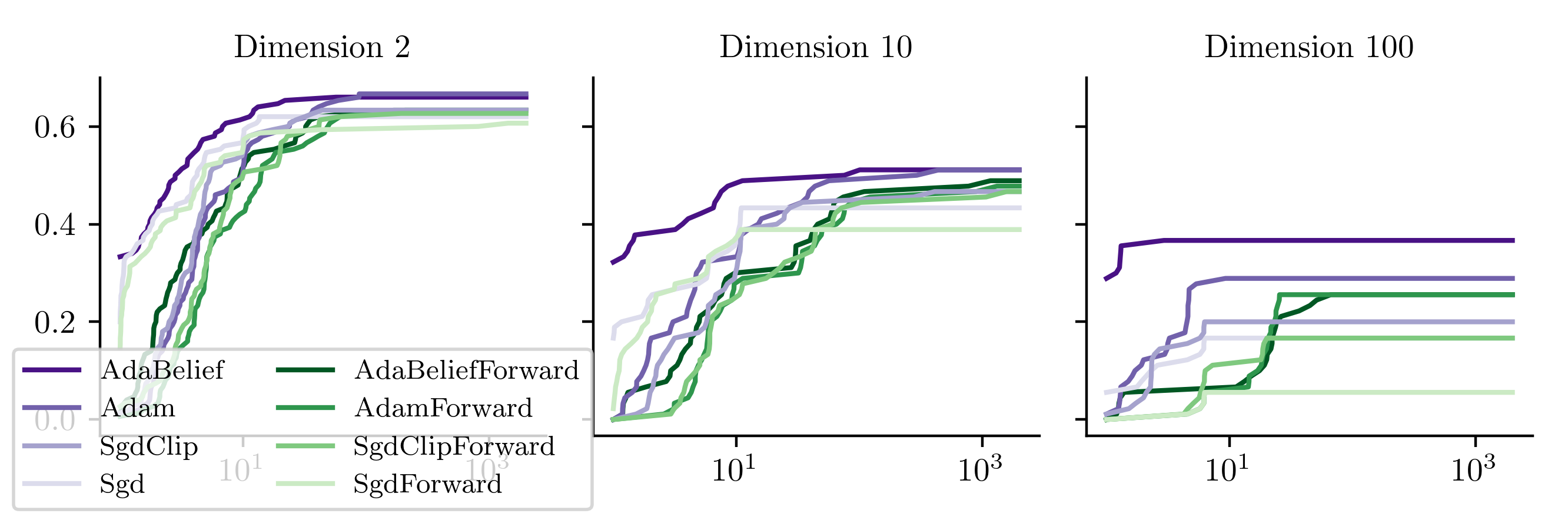}

\caption{Performance profiles across sets of problems with fixed dimensionality.}
\label{fig:performance-profiles-test-functions}
\end{figure*}

Performance profiles are convenient in that they display several information in a single graphic. However they treat all problems uniformly. In our case, we also would like to know how performance evolves with higher dimensionality. Thus, we provide in \autoref{fig:performance-profiles-test-functions} three profiles taken from problems with three different dimensions, $2$, $10$ and $100$.

Although we should be prudent with conclusions about 2-by-2 comparisons with performance profiles, we can observe the effect of higher dimensions, especially in the region $\tau \leq 10$. That is, in dimension $100$, any optimizer that uses forward gradients are almost always at least 10 times slower than the best optimizer that uses true gradient (which almost always is Adabelief). Note than on top of this, all optimizers converge less often in high dimensions (the SGD based optimizers seem to be the most impacted).

Moreover, we can see that for forward gradient based optimizers, the derivatives of the performance profile in $1$ gets flatter as the dimension increases $\rho'(1) \approx 0$. This means that not only are they almost never optimal (as $\rho(1) \approx 0$), but they also become poor approximators of the optimal solver in high dimensions. In contrast, Adam with true gradients has often good performance, albeit never optimal.

\section{Conclusion}
\label{ch:conclusion}

\paragraph{}Our work has exposed both theoretical and experimental shortcomings of forward gradients. Specifically, we identified high dimensional settings as an issue, and confirmed this result with extensive experiments.\\

On the theoretical side, we believe there are still many open roads for future work.

Most notably, we have not conducted any deep learning experiments, where \citeauthor{baydin_gradients_2022} reported encouraging results for forward gradients. We believe more experiments are needed to evaluate how well they compare to true gradients on more diverse architectures. More importantly, it remains mysterious to us as to why forward gradients would not struggle in such a high dimensional setting as neural networks are, and we trust that further work on this topic would bring a better theoretical understanding.

Alternatively, one could try to design optimizers specifically suited for forward gradients. For instance, the first moment in Adam, which acts as a predictor of what the gradient is going to be, could be used to bias the forward tangent distribution. Although our early experiments did not pan out, we are interested to explore this idea further.

\label{app:bibliography}
\raggedright
\printbibliography

\newpage
\appendix

\section{Appendix}
\label{ch:appendix}

\subsection*{Proof of \autoref{thm:minimal_deviation}}
\hypertarget{proof:minimal_deviation}{}

\begin{proof}
\begin{align*}
\esp\left[\left\|\nabla f (\bm{\theta}) - g(\bm{\theta})\right\|^2\right] &= \sum_i \bias(g_i(\bm{\theta})) + \var(g_i(\bm{\theta}))\\
&= \sum_i \esp[g_i(\bm{\theta})^2] - \esp[g_i(\bm{\theta})]^2\\
&= \sum_i \|\nabla f(\bm{\theta})\|^2 - \nabla f_i(\bm{\theta})^2\\
&= (n-1)\|\nabla f(\bm{\theta})\|^2
\end{align*}
\end{proof}

\subsection*{How far does a random walk go ?} \label{sec:random_walk_derivation}

\begin{lemma}[Divergence speed of a random walk]
Let $(X_i)$ be \iid centered Rademacher variables. Then \[\esp\left[|\sum_i^nX_i|\right] = \sqrt{\frac{2n}{\pi}} + O(\frac{1}{\sqrt{n}}) \]
\end{lemma}

\begin{proof}
We treat the case where $n=2N$ is even to alleviate notations. The case of odd $n$ is similar.
First we compute
\begin{align*}
\sum_{k=0}^N {2N \choose k} k &= 2N \sum_{k=0}^{N-1} {2N - 1 \choose k}\\
&= 2N4^{N-1}
\end{align*}
and
\begin{align*}
2\sum_{k=0}^{N-1}{2N \choose k} + {2N \choose N} &= 4^N\\
\Rightarrow \qquad \sum_{k=0}^{N-1}{2N \choose k} &= \frac{4^N - {2N \choose N}}{2}\\
\Rightarrow \qquad \sum_{k=0}^{N}{2N \choose k} &= \frac{4^N + {2N \choose N}}{2}
\end{align*}
Then
\begin{align*}
4^N\esp\left[|\sum_i^{2N} X_i|\right] &= \sum_{k=0}^{N-1}{2N \choose k}(2N - k - k) + \sum_{k=N+1}^{2N}{2N \choose k}(k - (2N - k))\\
&= 2\sum_{k=0}^{N}{2N \choose k}(2N - 2k) \\
&= 4N \frac{4^N + {2N \choose N}}{2} - 2N4^N\\
&= 2N {2N \choose N}
\end{align*}
Only remains now to estimate the asymptotic behaviour with Stirling:
\begin{align*}
\esp\left[|\sum_i^{2N} X_i|\right] &= \frac{2N}{4^N} {2N \choose N}\\
&\sim \frac{2N}{4^N} \frac{\left(\frac{2N}{e}\right)^{2N}\sqrt{4\pi N}}{\left(\frac{N}{e}\right)^{2N}2\pi N}\\
&\sim \frac{2N}{\sqrt{N\pi}} = \sqrt{\frac{2n}{\pi}}
\end{align*}
We can get the more precise result from the theorem by using one more term in the Stirling series, which we leave to the reader.
\[n! = \sqrt{2\pi n}(\frac{n}{e})^n\left(1 + O(\frac{1}{n}) \right)\]
\end{proof}
\end{document}